\documentclass{bmvc2k}
\pdfoutput=1

\title{Leveraging Geometry for Shape Estimation from a Single RGB Image}

\addauthor{Florian Langer}{fml35@cam.ac.uk}{1}
\addauthor{Ignas Budvytis}{ib255@cam.ac.uk}{1}
\addauthor{Roberto Cipolla}{rc10001@cam.ac.uk}{1}

\addinstitution{
 Department of Engineering\\
 University of Cambridge\\
 Cambridge, UK
}

\runninghead{Langer et al.}{Leveraging Geometry for Single Image Shape Estimation}

\usepackage{xcolor}
\newcommand*{\rom}[1]{\expandafter\@slowromancap\romannumeral #1@}
\begin{document}

\maketitle
\begin{abstract}
Predicting 3D shapes and poses of static objects from a single RGB image is an important research area in modern computer vision. Its applications range from augmented reality to robotics and digital content creation. Typically this task is performed through direct object shape and pose predictions \cite{meshrcnn,pixel2mesh,total_3D_understanding} which is inaccurate. A promising research direction \cite{mask2cad,Im2CAD,Location_Field_Descriptors} ensures meaningful shape predictions by retrieving CAD models from large scale databases \cite{chang2015shapenet,pix3D} and aligning them to the objects observed in the image. However, existing work \cite{mask2cad} does not take the object geometry into account, leading to inaccurate object pose predictions, especially for unseen objects. In this work we demonstrate how cross-domain keypoint matches from an RGB image to a rendered CAD model allow for more precise object pose predictions compared to ones obtained through direct predictions. We further show that keypoint matches can not only be used to estimate the pose of an object, but also to modify the shape of the object itself. This is important as the accuracy that can be achieved with object retrieval alone is inherently limited to the available CAD models. Allowing shape adaptation bridges the gap between the retrieved CAD model and the observed shape. 
We demonstrate our approach on the challenging Pix3D \cite{pix3D} dataset. The proposed geometric shape prediction improves the $ \mathrm{AP}^{\mathrm{mesh}} $ \cite{meshrcnn} over the state-of-the-art \cite{mask2cad} from 33.2 to 37.8 on seen objects and from 8.2 to 17.1 on unseen objects. Furthermore, we demonstrate more accurate shape predictions without closely matching CAD models when following the proposed shape adaptation. Code is publicly available at \url{https://github.com/florianlanger/leveraging_geometry_for_shape_estimation}.
\end{abstract}

\vspace{-0.5cm}
\section{Introduction}
\label{sec:intro}
\begin{figure}[t]
    \centering
    \includegraphics[width=1.0\linewidth]{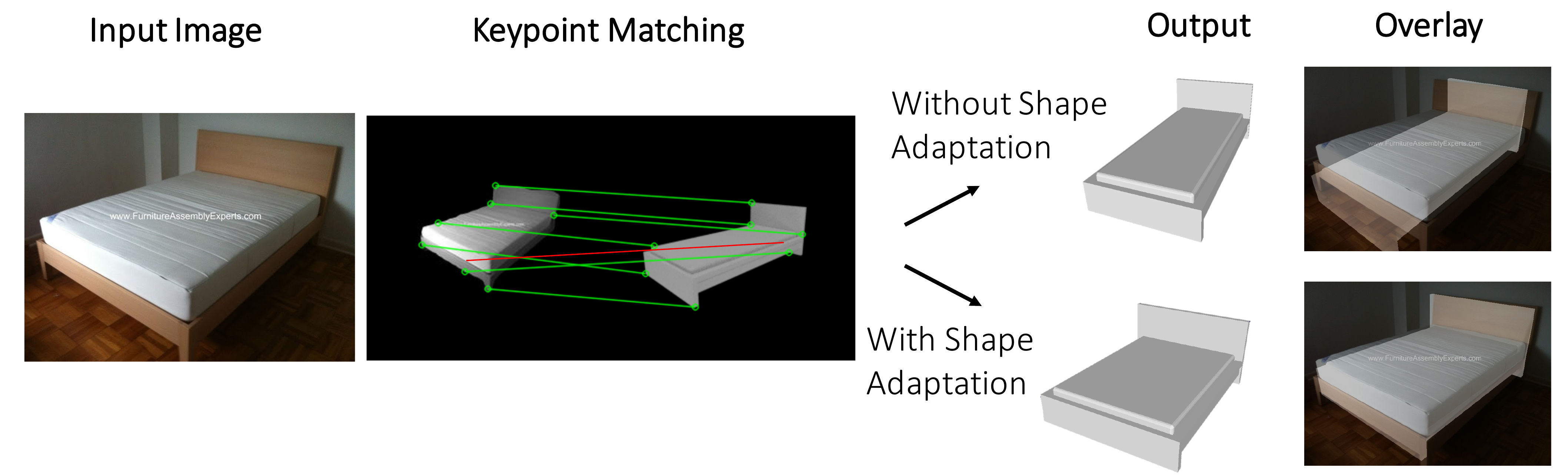}
    \caption{
    \textbf{Example result of our approach.} Given an input image we retrieve a CAD model rendering and perform key-point matching with the masked input image. Without shape adaptation the retrieved CAD model prediction is limited by the availability of similar CAD models in the database. When shape adaptation is performed target object shapes and their poses can be predicted precisely.} 
    \label{intro_fig}
\end{figure}
The past few years have seen rapid advances in object recognition in RGB images \cite{imagenet,maskrcnn,swin}. However, for many applications reasoning not in the 2-dimensional image but in the 3-dimensional world is crucial. One important task is 3D shape estimation from a single image, with applications ranging from robotics to augmented reality and digital content creation.
Current research on this task can be categorised into two different streams: \textit{direct shape prediction} and \textit{shape estimation via object retrieval}.\\ For direct object shape predictions different representations are used ranging from voxels \cite{voxels}, point clouds \cite{point_set,pointclouds}, meshes \cite{pixel2mesh,meshrcnn,TMN,total_3D_understanding}, packed spheres \cite{point_set}, binary space partitioning \cite{binary_planes}, convex polytopes \cite{polytopes}, signed distance fields \cite{sdf} to other implicit representations \cite{nerf}. Most of these approaches suffer either from a lack of precision \cite{voxels,point_set,pointclouds,pixel2mesh,meshrcnn,TMN,total_3D_understanding,polytopes} or lack applicability to a wide range of object classes \cite{binary_planes,sdf,nerf}.\\
Regardless of the representation chosen, directly predicting the shape and pose of an object from a single image is a difficult learning task. It is difficult as crucial information about the parts of an object without direct line of sight to the camera is absent at train and test time. The absence of complete shape information leads to an inherent ambiguity, as multiple consistent shape predictions may be possible; this ambiguity prevents the networks from learning effectively, causing them to perform class-averaged object shape predictions \cite{single_view_reconstruction_learn}.
A natural way to deal with the ill-posed nature of the problem lies in retrieving CAD models from existing large-scale databases \cite{chang2015shapenet}.
\cite{joint_embedding_space} and IM2CAD \cite{Im2CAD} show how CAD models and real images can be effectively encoded into a joint embedding space. At test time given a target RGB image multiple candidate CAD models can be retrieved. In order to estimate the object pose \cite{mask2cad} regress rotation parameters and use the ground truth $z$-coordinate for predicting object translation (see Section \ref{related_lit}).
While in this manner a neural network is able to store and interpolate between poses for objects that were seen during training, it fails to generalise to unseen objects. We avoid this problem by using a deep network to perform the simpler task of matching keypoints between the real image and the retrieved CAD model render \cite{keypoints_to_pose}, and using these matches as constraints to calculate the object pose analytically.
More importantly we show that we can use keypoint matches to modify the shape of the retrieved CAD model to better fit the object observed (see Figure~\ref{intro_fig}). We modify CAD model shapes by stretching them along the normal vector of 3D-planes. Stretching can be seen as a local operation which in contrast to a global scaling operation can modify proportions of objects within a single dimension (e.g. adjusting the height of the sitting area of a chair as shown in Figure \ref{fig_stretch_and_sil}). Currently our approach uses stretching along the three principal object axis. However, in the future predicting additional stretch planes and limiting stretch extents to variable 3D-boxes will allow for even more fine-grained shape adaptations.\\
We evaluate our approach on the Pix3D \cite{pix3D} dataset. When combining object retrieval with a geometric pose prediction we outperform existing work \cite{meshrcnn,mask2cad} on splits containing both seen and unseen CAD models at train time (see Table \ref{S1_table}). We evaluate the proposed object adaptation on a range of adaptation experiments. Here we observe that dynamic fitting improves the shape predictions when no access to correct CAD models is given at test time and retrieved models have to be adapted (see Figure \ref{fig_shape}).

\vspace{-0.4cm}
\section{Related Work} \label{related_lit}
\vspace{-0.2cm}
Related literature on shape estimation for static objects from single images can be categorised into direct shape prediction methods and retrieval based methods.\\
\textbf{Direct Prediction Methods}.
Current direct prediction methods differ greatly in the choice of object representations that are used. In the \textit{voxel} representation \cite{voxels} the 3D world is discretised into cubes and object shapes are encoded as binary occupancies of these cubes. This allows the shape prediction task to be formulated as a binary classification task which is usually simpler to learn compared to a regression task. However, the voxel representation suffers from an inherent trade off between accuracy and storage space due to the cubic scaling. \textit{Meshes} \cite{meshrcnn} alleviate the storage-accuracy trade-off by encapsulating information about the 2D object surface rather than its 3D volume. An object is represented as a set of vertices which are interconnected to form (usually triangular) mesh faces. 
The difficulties associated with meshes are twofold; predicting the correct object topology and predicting precise vertex positions. 
Early work \cite{pixel2mesh} simply deformed an original ellipsoid mesh and was therefore not able to predict shapes with complex topologies. 
In contrast \cite{meshrcnn} first predict a rough shape estimate in the voxel representation. This serves as the initialisation for the topology of the shape and is subsequently refined as a mesh. Recently Topology Modification Networks \cite{TMN} aim to directly predict varying object topologies. Other works decompose objects into a set of \textit{convex polytopes} \cite{polytopes} or assemble 3D shapes from \textit{geometric primitives} \cite{primitives}. Besides these explicit object encodings other implicit representations exist for defining the object boundary. \cite{binary_planes} approximate an object by predicting a set of \textit{3D planes} and the corresponding side on which the object lies on. Given a 3D point \textit{signed distance fields} \cite{sdf} predict the distance of the point from the object surface and whether it resides within or outside of the object. \textit{Neural radiance fields} \cite{nerf} encode information not just about the object shape but also its texture. While these implicit representations are promising research directions, they require vast amounts of training and have not been shown yet to work on a large number of different shapes in realistic settings.\\
\textbf{Object Retrieval}.
Almost all existing work on object retrieval constructs an embedding space of CAD model renderings. At test time a real image is embedded into this space and its nearest neighbours are retrieved. While early work \cite{seeing_chairs} construct and map into this space based on the Histogram-of-Gradient \cite{hog} (HoG) descriptor, \cite{joint_embedding_space} learn real image embeddings using a convolutional neural network. Instead of constructing the embedding space from HoG-descriptors \cite{Im2CAD} learns a \textit{joint embedding space} between CAD model renderings and real images. In order to align retrieved CAD models with the objects in the image \cite{Im2CAD} follows a \textit{Render-and-Compare} approach in which they optimise for an object pose by iteratively comparing the re-rendered CAD model embedding to the original image embedding and update the pose accordingly. While \cite{Im2CAD} can successfully predict rough room layouts and retrieve similar CAD models to the ones observed, their approach is significantly less accurate in terms of object retrieval and pose prediction than ours. The work most similar to ours is Mask2CAD \cite{mask2cad}. In contrast to our method \cite{mask2cad} estimates the object pose by regressing the rotation parameters as well as the offset between the reprojected object center and the 2D bounding box centers. Using the ground truth $z$-coordinate \cite{mask2cad} can estimate the object pose. Unlike our system \cite{mask2cad} is limited to pure shape retrieval as it can not perform shape adaptations. Patch2CAD \cite{patch2cad} is similar to \cite{mask2cad} but instead of retrieving a single CAD model based on the entire image, \cite{patch2cad} retrieves many CAD models for different image regions and perform majority voting to obtain the final prediction. \cite{Location_Field_Descriptors} learn an embedding space and estimate poses using a Location Field Descriptor containing the estimated 3D object coordinate for every pixel. However, their method struggles to predict correct location fields from which the pose can be accurately computed and do not work for occluded objects. Recently \cite{vid2cad} have extended \cite{mask2cad} to temporally combine predictions from individual video frames into more accurate CAD model and pose predictions. We differ from all existing work on object retrieval in the usage of keypoint matches for precise pose and shape estimation.
\vspace{-0.4cm}
\section{Method}
\vspace{-0.2cm}
Our method consists of four steps: (i) object detection and instance segmentation, (ii) CAD model retrieval, (iii) keypoint matching and finally, (iv) pose and shape optimisation.\\
\begin{figure*}[t]
    \centering
    \includegraphics[width=1.0\linewidth]{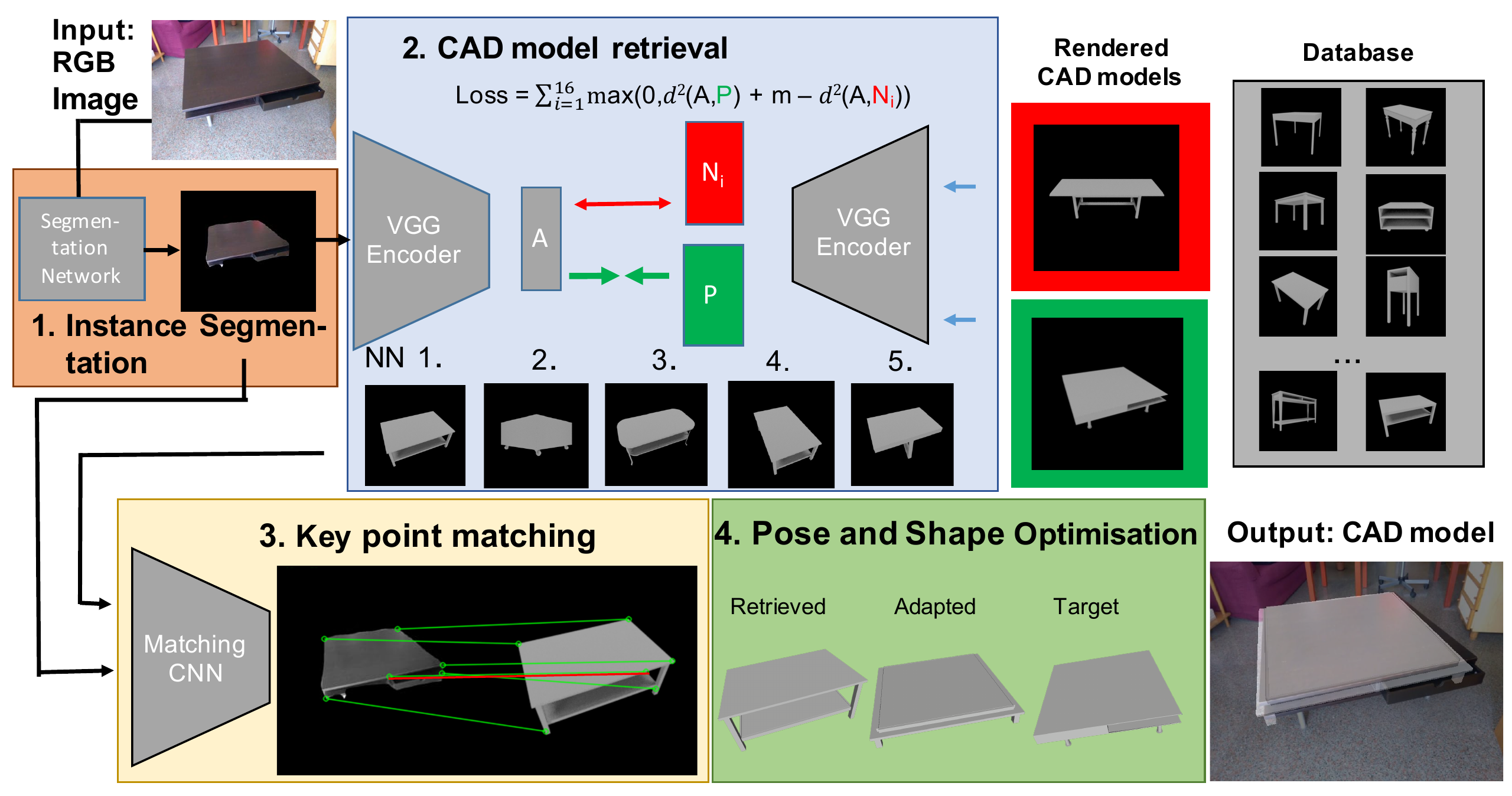}
    \caption{\textbf{Method:} Given an RGB image we perform object detection and instance segmentation (step 1). We then retrieve a set (e.g. 10) of the nearest neighbour CAD model renderings (step 2) and perform keypoint matching between the retrieved image and the rendered image using SuperPoint~\cite{superpoint} (step 3). The keypoint matches are subsequently used to jointly optimise over the shape and pose of the object (step 4).}
    \label{fig_method}
\end{figure*}
\noindent
\textbf{Object Detection and Instance Segmentation.}
For object detection and instance segmentation we train a Swin-Transformer \cite{swin} network on the Pix3D \cite{pix3D} dataset. During training we employ standard image augmentation techniques including random crops, scaling, rotations, horizontal flipping, and random brightness and contrast adjustments. We also report results on segmentation masks obtained using Mask R-CNN \cite{maskrcnn} trained by \cite{meshrcnn} as our approach is sensitive to the quality of segmentation predictions. The Swin-Transformer network provides more accurate segmentations of objects with hard edges (e.g. chairs, tables or wardrobes), while Mask R-CNN is superior on objects from categories with soft edges such as beds and sofas.\\
\textbf{Learning a Joint Embedding Space.}
In order to map CAD models and masked RGB images into a joint embedding space, we first render a CAD model in regular intervals and represent it as a collection of these renderings \cite{lightfield_descriptor}. This step is important as it bridges the domain gap from CAD models to RGB images and increases the similarity of the two inputs therefore simplifying the matching task. We use a single VGG \cite{vgg} encoder for encoding both real masked RGB images and rendered inputs. This encoder is trained using a triplet-loss \cite{triplet_loss}:
\begin{equation}
\label{eq_triplet_loss}
    \mathcal{L} = \sum_{i=1}^{16} \mathrm{max} (0 , d^2(A,P) - d^2(A,N_i)+m) \, .
\end{equation}

Here $A$ is the encoding of an anchor RGB image, $P$ is the encoding of the positive example and $N_i$ is the encoding of a negative example. Only the rendering of the corresponding ground truth CAD model in the most similar orientation to the object is considered to be the positive example. A random subset of 16 renderings of different CAD models are used as negative examples $N_i$. Finally, $m$ is the margin and $d(x,y)$ is the Euclidean distance over the $128$-dimensional encodings.
As in \cite{mask2cad} we employ hard example mining during training. Only renderings of CAD models of the same category as the query image are considered for hard-negative mining. 
Unlike \cite{mask2cad}, no hard example mining is applied to positive examples. Doing hard positive mining forces the network to match very different views, often not sharing any features with each other, therefore necessarily leading to poor performance via over-fitting. Instead, as mentioned above, the CAD model rendering in the most similar orientation is used as a positive example.
At test time we use an embedding of a masked RGB image and retrieve the nearest neighbour CAD model renderings which are then passed on for keypoint matching and joint pose and shape estimation.

\noindent
\textbf{Key-Point Matching.}
In order to precisely estimate poses, keypoint matching is performed between the masked RGB image and its retrieved CAD model rendering~\cite{keypoints_to_pose}. Finding and matching valid keypoints across different domains (RGB and rendered image) may seem initially a difficult task due to the contrast of cleanly (sharp boundaries, perfect segmentation masks, no occlusions) rendered images and varying appearances of objects in RGB images (variety of textures, lighting conditions and imperfect segmentation masks).
However, we found a SuperPoint \cite{superpoint} keypoint detection and matching network to be well suited for this task. Its robustness stems from the fact that it was initially trained to detect corners of triangles, quadrilaterals, lines, cubes, checkerboards and stars in synthetic images which was followed by fine-tuning on real images, hence performing well on both domains. Crucially man-made furniture objects contain many of the aforementioned primitives.
We used the trained off-the-shelf network implementation from~\cite{superpoint} to avoid over-fitting as Pix3D \cite{pix3D} is considerably smaller than typical datasets used for keypoint detector training. SuperPoint \cite{superpoint} returns approximately $25$ keypoints (each described with a 256-D vector) on average per real RGB image ($15$ on rendered image) using a default confidence threshold of $0.015$. Nearest neighbour matches are found for each keypoint using the L2-distance. Cross-checking is used to eliminate one-sided matches producing on average $12$ matches per RGB and rendered image pair.

\noindent
\textbf{Pose Estimation.}
\begin{figure*}[t]
    \centering
    \includegraphics[width=1.0\linewidth]{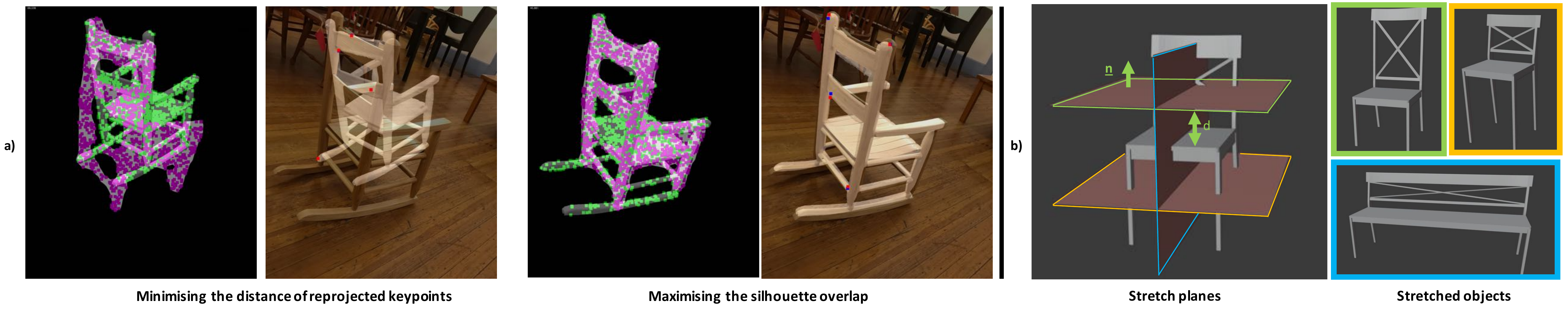}
    \caption{Left: \textbf{Pose Selection.} Green points are reprojected points from the CAD model under the current pose estimate. Purple points are sampled inside the predicted segmentation mask. Selecting a pose based on the approximated silhouette overlap yields precise poses (top) while using the minimum distance of the reprojected keypoints selects inaccurate poses (bottom). Right: \textbf{Stretching Procedure.} A CAD model can be stretched along different stretch planes to generate new shapes. The plane normal $\mathbf{n}$ and the distance $d$ of the plane from the object center are highlighted for one of the stretch planes. }
    \label{fig_stretch_and_sil}
\end{figure*}
The keypoint matches provide correspondences between real image pixel coordinates (2D) and 3D world coordinates in CAD model space. This allows us to formulate the pose estimation as a PnP-problem. Since the obtained matches are often noisy (only 4-5 matches out of 12 are correct on average) we estimate poses for all available quadruplets\footnote{On average we obtain 12 keypoint matches per image leading to $12 \times 11 \times 10 \times 9 / ( 4 \times 3 \times 2 \times 1) = 495$ possible quadruplets and poses.} of matches using the UPnP~\cite{upnp} algorithm. Instead of using an inlier scoring as is typically done in robust pose estimation, we select a final pose from the computed poses by approximating the Intersection-over-Union (IoU) silhouette overlap of the reprojected CAD model and the predicted segmentation mask. For this purpose we sample 1000 points from the CAD model and reproject them into the RGB image. Additionally we sample 1000 points from within the predicted segmentation mask. For each reprojected point we compute the distance to the closest points sampled from within the predicted segmentation mask and vice versa (see Figure \ref{fig_stretch_and_sil} on the left).
For each object pose the average of the largest 
$20\%$ of pairwise distances is computed and the pose
with the smallest average is selected as the final pose. We found that selecting the final pose based on the average of all pairwise distances was less robust in pose estimation (see supplementary material).

\noindent
\textbf{Pose and Shape Estimation.} In order to perform joint pose and shape adaptation the standard PnP-formulation is not sufficient. In particular, to enable shape adaptation, we reparametrise previously fixed object coordinates $x$ (now $x_{stretch}$) in the following way. We allow stretching of the object by an amount $\tau_i$ along the normal $\mathbf{n}_i$ of a plane $P_i$ defined by $\mathbf{n}_i \cdot \mathbf{x}_i = d_i $. Here $d_i$ is the distance of the plane $P_i$ from the object center (see Figure \ref{fig_stretch_and_sil} on the right). Repeating this stretching for $N_{\mathrm{planes}}$ orthogonal planes the stretched world coordinates become
\begin{equation}
\mathbf{x}_{\mathrm{stretch}} = \mathbf{x} + \sum_{i=1}^{N_{\mathrm{planes}}}s_i \cdot \frac{\tau_i}{2} \cdot \mathbf{n}_i \quad \quad \mathrm{where} \, \, s_i =  \begin{cases}
    1,& \text{if } \mathbf{x} \cdot \mathbf{n}_i \geq d_i\\
    0,& \text{if } \mathbf{x} \cdot \mathbf{n}_i = d_i\\
    -1,& \text{if } \mathbf{x} \cdot \mathbf{n}_i \leq d_i
\end{cases} \, .
\end{equation}
From the stretched world coordinates one obtains reprojected pixel $\mathbf{v} \in \mathcal{R}^2$ under the perspective camera model $(s \, v_x,s \, v_y,s) = \mathbf{K} \left[\mathbf{R} | \mathbf{T} \right] \left[ \mathbf{x}_{\mathrm{stretch}} , 1 \right] ^T$ for camera calibration matrix $\mathbf{K} \in \mathcal{R}^{3 \times 3}$, rotation matrix $\mathbf{R} \in \mathcal{R}^{3 \times 3}$ and translation vector $\mathbf{T} \in \mathcal{R}^{3 \times 1}$. The rotation matrix is parameterised in terms of Euler angles $\mathbf{\theta} \in \mathcal{R}^3$. For known camera intrinsics the objective function to be minimised is therefore $f = \sum_{j=1}^{N_{\mathrm{matches}}} (\mathbf{u}_j - \mathbf{v}_j)^2 $ with respect to $(\mathbf{\theta},\mathbf{T},\mathbf{\tau})$ where $\mathbf{u}_j$ are pixel coordinates of the j-th match in the RGB image. L-BFGS \cite{lbfgs} minimiser, instead of UPnP is used for the minimisation of this non-linear objective function. It is initialised with the original pose of the retrieved CAD model and no stretching $\mathbf{\tau} = 0$. Note that in this case we sample sets of 6 matches instead of 4 to cover enough degrees of freedom for 3 deformations. 


\vspace{-0.4cm}
\section{Experimental Setup}
\vspace{-0.2cm}
This section briefly describes the Pix3D \cite{pix3D} dataset that was used for training and evaluation, the $\mathrm{AP}^{mesh}$ metric we adopted for evaluation as well as the hyperparameters chosen.\\
\textbf{Pix3D Dataset.}
The Pix3D \cite{pix3D} dataset consists of 10,069 RGB images annotated with aligned 3D CAD models (one per image). There are a total of 395 different CAD models from 9 categories (chair, sofa, table, bed, desk, bookcase, wardrobe, tool and miscellaneous). For our experiments we consider two splits originally proposed by \cite{meshrcnn}.\\
\textbf{S1 split.} The $S1$ split randomly splits the 10,069 images into 7539 train images and 2530 test images. In this split all CAD models are seen during training and the challenge is to retrieve and align the correct CAD model from images containing different scenes where (possibly occluded) objects appear with new textures under varying lighting conditions.\\
\textbf{S2 split.} Under the S2 split train and test images are split such that the CAD models that have to be retrieved at test time were unseen during training. This split is more difficult as it prohibits the embedding network to simply remember CAD models and truly tests its ability to learn meaningful embeddings.\\
\textbf{Evaluation metric.}
We adopt the commonly used $\mathrm{AP}^{mesh}$ metric \cite{meshrcnn} for evaluating the retrieved object shapes. Following the standard COCO \cite{coco} object detection protocol of AP50-AP95 (denoted AP), we average over 10 IoU thresholds ranging from 0.50 to 0.95 in 0.05 intervals. For a given threshold the $\mathrm{AP}^{mesh}$ score is defined as the mean area under the per-category precision-recall curve where a shape prediction is considered a true-positive if its predicted category label is correct, it is not a duplicate detection, and its $\mathrm{F1}^\tau$ is greater than the IoU threshold. For a given predicted shape the $\mathrm{F1}^\tau$ score is the harmonic mean of the fraction of predicted points within $\tau$ of a ground-truth point and the fraction of ground-truth points within $\tau$ of a predicted point. We follow \cite{meshrcnn,mask2cad} in choosing $\tau = 0.3$. For fair comparison across different object sizes we rescale all objects such that the longest edge of the ground truth model's bounding box has length 10 before computing the $\mathrm{F1}$ score.\\
\textbf{Hyperparameter settings}.
CAD models are rendered at $16$ regularly sampled azimuthal angles spanning $360^{\circ}$ and $4$ different elevation angles between $0^{\circ}$ and $45^{\circ}$. The VGG encoder is trained with a batchsize of $8$ real images as each example requires $16$ negative anchors and one positive anchor leading to a total of $144$ images per batch. We use a learning rate of $ 2 \times 10^{-6}$ and set the margin of the triplet-loss in Equation \ref{eq_triplet_loss} to $m=0.1$. We use off-the-shelf object detection and segmentation networks \cite{swin,maskrcnn} as well as keypoint matching network - SuperPoint \cite{superpoint}.
\vspace{-0.4cm}
\section{Experimental Results}
\vspace{-0.2cm}
This section showcases our experimental results. We compare against Mesh-RCNN \cite{meshrcnn} and Pixel2Mesh \cite{pixel2mesh} (specifically the reimplementation by \cite{meshrcnn} which outperforms the original \cite{pixel2mesh}) as well as Mask2CAD \cite{mask2cad} and Patch2CAD \cite{patch2cad}. Section \ref{subsec_access_gt_Cad_models} shows our results on the Pix3D\cite{pix3D} S1 and S2 split when access to the correct CAD models is provided at test time. In Section \ref{subsec_no_access_gt_Cad_models} we evaluate the proposed stretching on modified versions of the original CAD models as well as when available CAD models have to be adapted to match entirely different ones.

\begin{figure*}[t]
    \centering
    \includegraphics[width=1.0\linewidth]{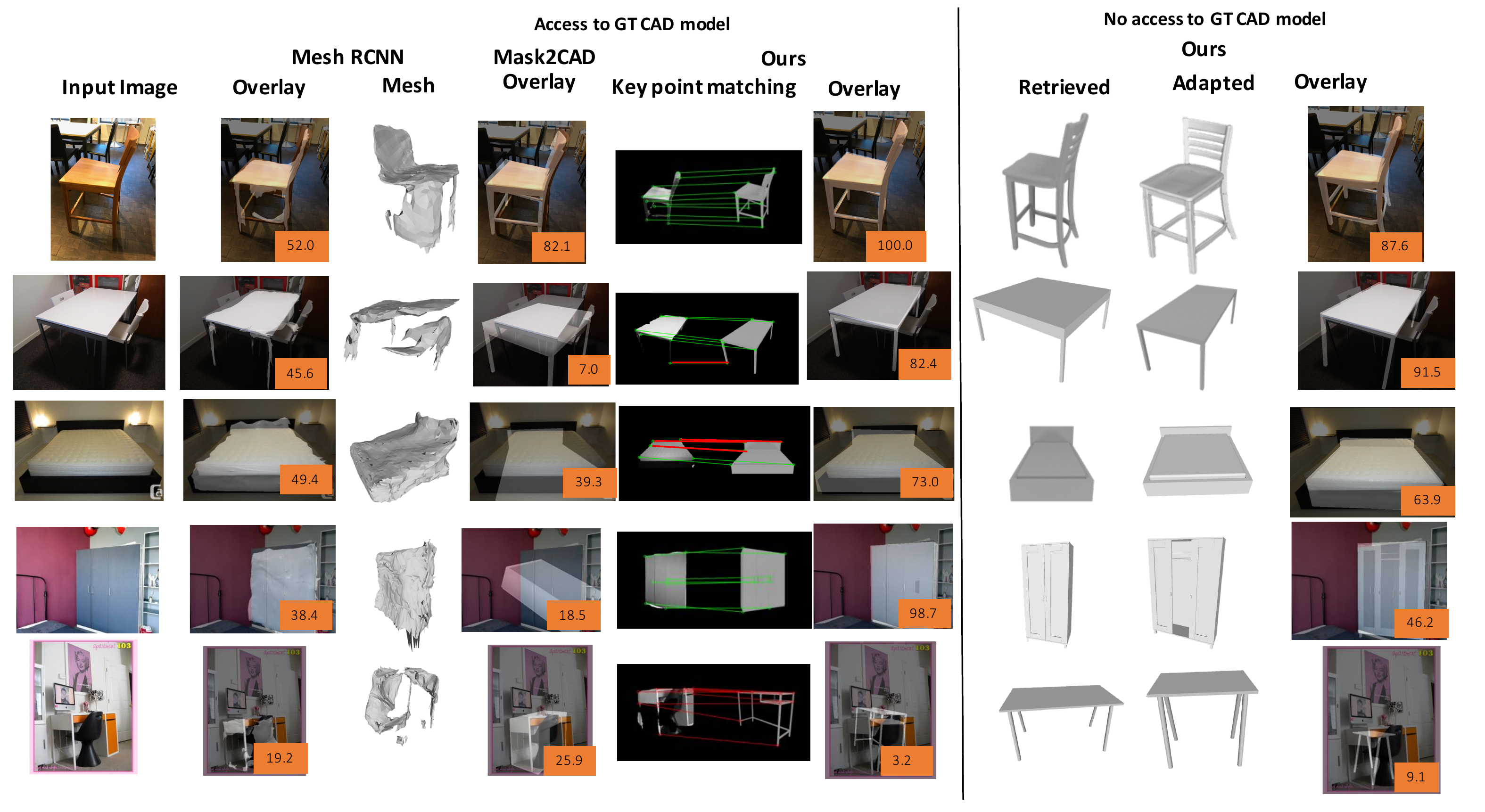}
    \caption{\textbf{Qualitative comparison for predictions on the S2 split:} The left side shows results when access to the correct CAD model is given at test time, but which were unseen at train time. The right side shows the case when no access to correct CAD models is given and retrieved CAD models have to be adapted dynamically. Overlaid numbers are the $\mathrm{F1}$ score. In general the comparison shows that a geometric approach allows for very precise pose estimation whereas the direct prediction method of Mask2CAD \cite{mask2cad} is less precise. Note that the qualitative results are from our reimplementation of \cite{mask2cad} as neither the code nor the predictions are publicly available. In comparison to CAD model retrieval direct mesh predictions \cite{meshrcnn} are very imprecise, often failing to predict the correct topology and performing particularly poorly on the backside of objects. Row 4 shows the sensitivity of the $\mathrm{F1}^{0.3}$ score. Despite an appropriate object retrieval and very good shape adaptation, slight imprecision in the alignments lead to a low F1 score. Finally row 5 shows a failure case of ours where poor segmentation leads to a wrong shape retrieval and correspondingly false keypoint matches resulting in a bad final pose and shape.}
    \label{fig_qual_no_stretch}
\end{figure*}
\vspace{-0.3 cm}
\subsection{Geometry-Based Shape and Pose Predictions are very Precise}
\label{subsec_access_gt_Cad_models}

Table \ref{S1_table} shows the $\mathrm{AP}^{mesh}$ we obtain on the S1 and S2 splits of Pix3D, originally proposed by Mesh R-CNN \cite{meshrcnn}. For a fair comparison to Mask2CAD \cite{mask2cad}, Patch2CAD \cite{patch2cad}, Mesh R-CNN \cite{meshrcnn} and Pixel2Mesh \cite{pixel2mesh} we use the ground truth $z$-coordinate for the final pose. While our approach does not require the use of the ground truth $z$-coordinate (unlike \cite{mask2cad,patch2cad}) it improves our performance as the low F1 score threshold is very sensitive even to small displacements in the $z$ direction arising from slight inaccuracies in the keypoint matches.

\noindent
\textbf{Seen Objects.} Results on the S1 split show that particularly on seen objects, CAD model retrieval is more precise than direct predictions. For the high $ \mathrm{AP}^{\mathrm{mesh}} $ regions (i.e. AP and AP75) retrieval-based methods outperform direct prediction methods by a large margin. On the AP50 score Mesh-RCNN \cite{meshrcnn} performs similar to the retrieval-based methods as the AP50 score only evaluates if at least $50\%$ of a sampled object is predicted correctly and Mesh-RCNN \cite{meshrcnn} is generally able to predict the object side in direct sight of the camera. In terms of the category average we outperform the best competitor \cite{mask2cad} (37.8 vs 33.2). Mask2CAD \cite{mask2cad} performs well on large planar objects such as bookcases or wardrobes, whereas we have very strong performance on high fidelity objects, such as chairs, allowing for numerous keypoint matches.
When using ground truth masks compared to predicted masks we observe a very large performance gain (37.8 vs 33.2). This observation is crucial as it shows the potential of our approach with improved segmentation masks. While competing approaches will also benefit from better segmentation, having accurate instance masks for the pose prediction is not as an integral part in their pipeline as it is for our keypoint matching and does therefore not benefit them as much.

\noindent
\textbf{Unseen Objects.}
We show qualitative comparisons on the S2 split for \cite{mask2cad} and \cite{meshrcnn} in Figure \ref{fig_qual_no_stretch}. Quantitatively, we outperform all competitors on all metrics. 
We perform significantly better than \cite{mask2cad} (17.1 compared to 8.2 on the class average) because of the geometric pose predictions. \cite{mask2cad} struggles to predict poses of unseen objects whereas our pose prediction relies on analytically computing poses from correspondences which generalises a lot better to unseen shapes. Note that Mask2CAD \cite{mask2cad} performs well on sofas, not because it is able to retrieve an unseen sofa but because for every sofa in the S2 split there is a very good fitting sofa among the seen sofas, allowing to simply retrieve that for a good performance (see the supplementary material). 




\begin{table}[t]
    \centering
    \includegraphics[width=1.0\linewidth]{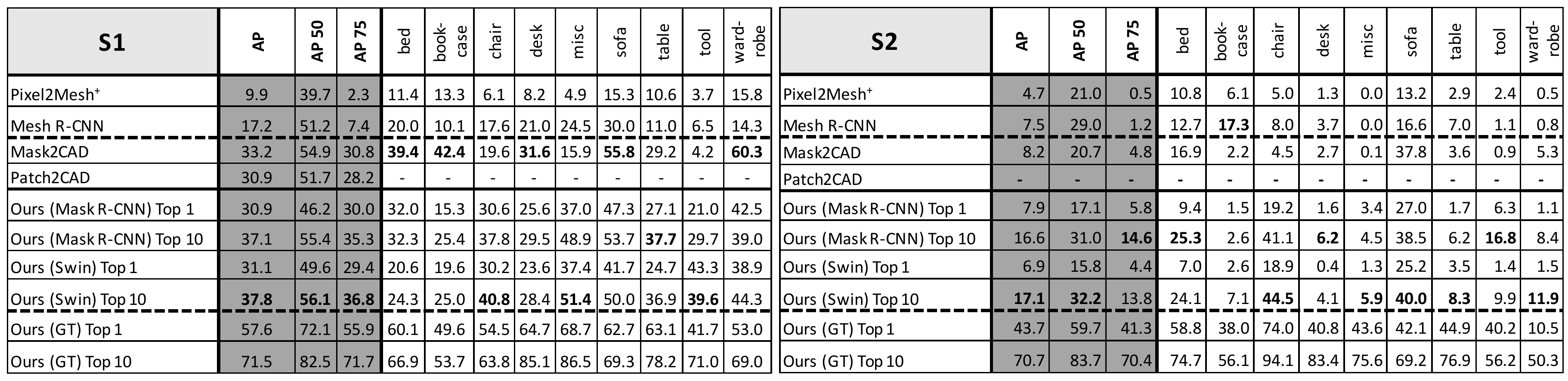}
    \caption{\textbf{Quantitative results on the S1 and S2 split} for seen and unseen objects on the Pix3D dataset. Brackets indicate the segmentation masks that were used. Bold numbers are the maximum values excluding experiments that were run with ground truth (GT) masks.}
    \label{S1_table}
\end{table}

\begin{figure*}[!]
    \centering
    \includegraphics[width=1.0\linewidth]{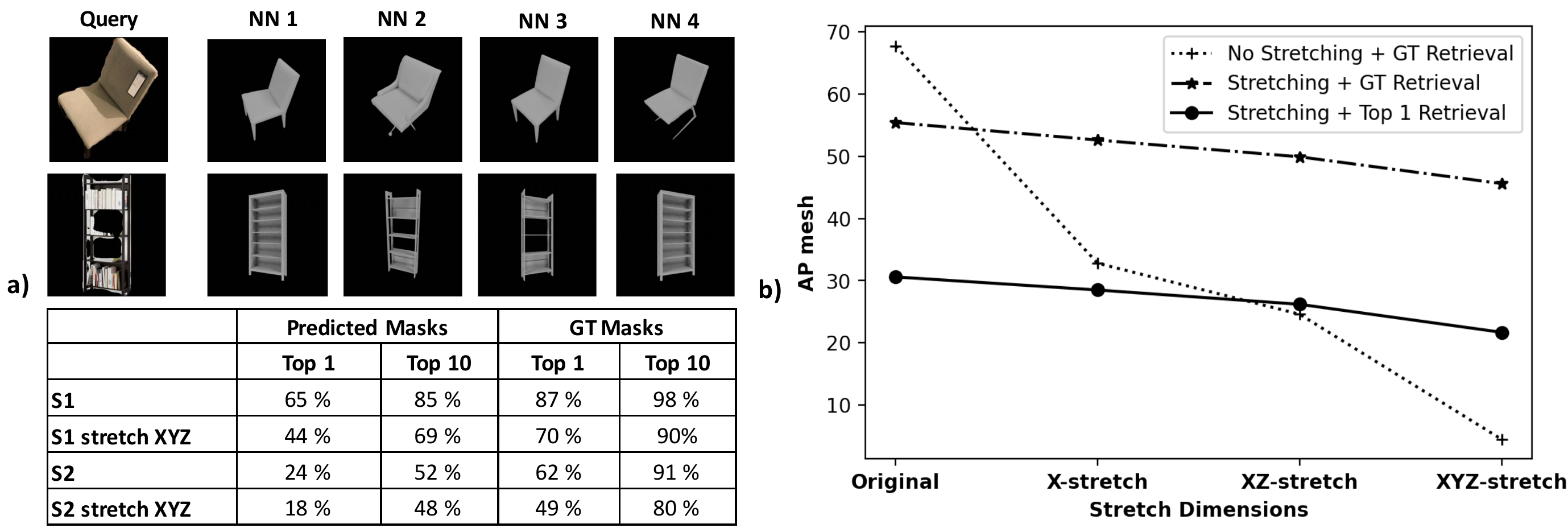} 
    \caption{a) \textbf{Retrieval accuracy} for selected CAD model splits. When considering the top 10 nearest neighbours the retrieval network is able to return completely unseen CAD models in over 50\% of cases. Note that different renderings of the same CAD model are considered as different nearest neighbours. b) \textbf{Ablation experiments}  on the proposed object stretching with ground truth masks. We plot the average $ \mathrm{AP}^{\mathrm{mesh}} $ score as a function of increasing shape deformations of S2 CAD models. On the left no deformations were performed while on the right objects were stretched along the $x$,$y$ and $z$ direction. With increasing deformation simple object retrieval quickly becomes inaccurate, while the proposed stretching is able to maintain a high accuracy.}
    \label{table_retrieval_accuracies}
\end{figure*}
\vspace{-0.4cm}
\subsection{Shape and Pose Estimation using Stretchable CAD Models}
\begin{figure*}[!]

    \centering
    \includegraphics[width=1.0\linewidth]{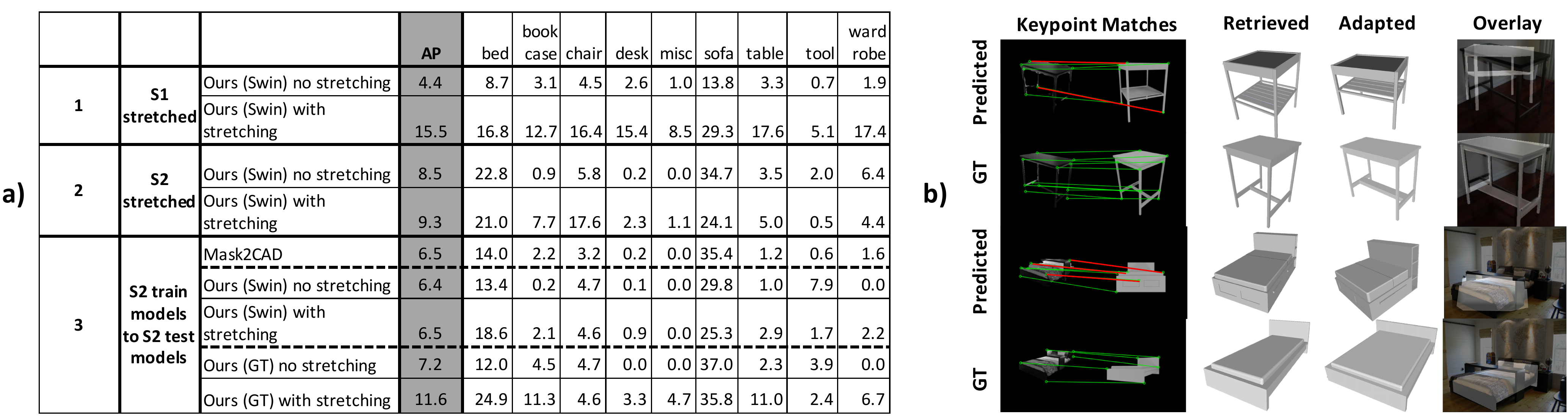}
    \caption{a) \textbf{Quantitative results on Pix3D} when no access to correct models is provided at test time. For the first two main rows (containing two sub-rows each) CAD models in the database are randomly stretched along all 3 principal directions and our method has to recover the original shape. For the third main row S2 test CAD models have to be estimated when the retrieval network has no access to the correct models and differing S2 train CAD models have to be adapted. b) \textbf{Visual comparison} of shape prediction with predicted (row one and three) and ground truth (row two and four) segmentation masks.}
    \label{fig_shape}
\end{figure*}

\vspace{-0.1cm}
\label{subsec_no_access_gt_Cad_models}
For realistic settings a given object will not have a perfect match among the available CAD models and a retrieved CAD model will require adaptation. We evaluate our adaptation approach on three different settings: \textit{stretched S1 models}, \textit{stretched S2 models} and predicting \textit{S2 test models with S2 train models} (see Figure \ref{fig_shape}). For the first two experiments the databases of S1 and S2 models are modified by applying random stretching in the $x$, $y$ and $z$ direction with planes passing through the center of the object. Stretch factors $\tau_i$ in each direction are obtained by multiplying a random number from a uniform distribution on the interval [-0.2, 0.3] by the side length of the bounding box of the object in the relevant direction.
Experiments on \textit{stretched S1 models} demonstrate that shape adaptation substantially improves the shape predictions (15.5 vs. 4.4, see Figure \ref{fig_shape}). For \textit{stretched S2 models} we can observe significant improvements for certain classes such as bookcases, chairs or tables. However, the overall accuracy gain is smaller (9.3 vs. 8.5)  and some classes,  
 most notably sofas, show worse performance. As explained above very accurate shape models already exist for sofa objects in the non-stretched part of the database. Hence, stretching from sometimes poor segmentation masks and corresponding keypoint matches can deteriorate the estimated shapes. To further investigate the performance of the system when enabling shape deformations, we perturb the models in the database with progressively larger perturbations, and compare the predictions generated when stretching is allowed and disallowed (see Figure \ref{table_retrieval_accuracies} b).
 As CAD model perturbations are increased, the proposed shape adaptation method can maintain a high accuracy while estimating the poses and of retrieved objects without using shape adaptation quickly becomes inaccurate.
Finally, when estimating shapes \textit{on S2 test images while only allowing to retrieve from the S2 train model set} (see Figure \ref{fig_shape}) we observe that similar to the experiments on the  \textit{stretched S2 models} we show improvements only for some classes (e.g. beds). This is due 
to the poor segmentation quality which prevents the matching network from successfully establishing
a sufficient number of correspondences. When ground truth masks are used instead, significant improvements for almost all categories are observed.

\vspace{-0.4cm}
\section{Conclusion}
\vspace{-0.2cm}
In this work we propose to leverage geometric constraints for precisely estimating 3D object shapes from retrieved CAD models. We demonstrate that our approach is more accurate than direct pose prediction \cite{meshrcnn,pixel2mesh} and the image retrieval-based approaches \cite{mask2cad,patch2cad}. Further, we show that by allowing object stretching we can modify retrieved CAD models to better fit the observed shapes. We believe that adapting retrieved CAD models is a promising avenue for future research as it combines the reliability of object retrieval with the expressiveness of generative approaches.

\section*{Acknowledgments}
This work was supported by the Engineering and Physical Sciences Research Council\\
{[RG105266]}.

\bibliography{bmvc_final}

\end{document}


\maketitle

\section{Pretrained SuperPoint Network}
We use a SuperPoint \cite{superpoint} network for detecting keypoint matches between real RGB images and rendered CAD models. We use an off-the-shelf version of SuperPoint \cite{superpoint} with pre-trained weights and confidence threshold of $0.015$. We include a brief summary of the training process below. More information can be found in \cite{superpoint}. SuperPoint \cite{superpoint} was initially trained to detect corners of triangles,
quadrilaterals, lines, cubes, checkerboards and stars in synthetic images. The initial baseline detector that was trained on synthetic images was then applied to real images. By performing a set of random homographic adaptations (consisting of random translation, scaling, in-plane rotations and perspective distortions) interest points are detected in the adapted images. These are then accumulated in the original image and serve as pseudo-ground truth interest points which the network is trained to detect. The extensive training for detecting corners both in real and in synthetic images makes SuperPoint \cite{superpoint} suitable for our cross-domain matching task.

\section{Ablations}
\label{sec_ablations}
We perform a series of ablation experiments (see  Table \ref{fig_all_ablation}).

\begin{table}[t]
    \centering
    \includegraphics[width=\linewidth]{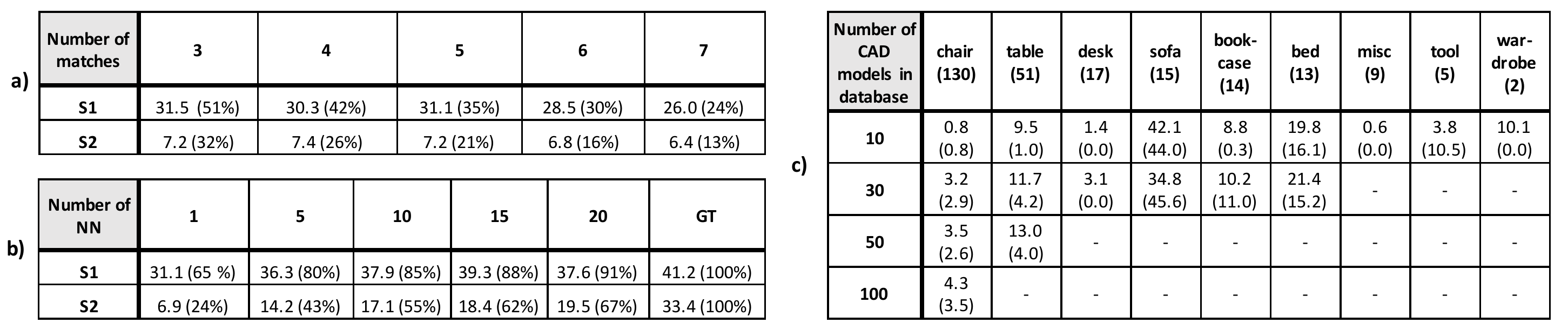}
    \caption{Numbers displayed are $\mathrm{AP}^{\mathrm{mesh}}$ scores averaged across all classes. a) The ablation for the number of matches was performed with top 1 retrieval results and Swin-Transformer \cite{swin} segmentation masks. Numbers in brackets indicate how many of the test images have at least $n$ correct matches. Here a match is considered correct if the reprojection error of the reprojected keypoint under the ground truth pose is less than 5 pixel. b) For the ablation on the number of nearest neighbours numbers in brackets indicate the retrieval accuracy i.e. for how many of the images the correct object is among the top $n$ retrieved. GT indicates ground truth retrieval of the correct object in the most similar pose. c) When investigating the adaptation performance for variable size datasets ground truth masks are used. Numbers in brackets indicate the $\mathrm{AP}^{\mathrm{mesh}}$ score when no shape adaptation is performed and numbers next to category names indicate the total number of CAD models in the $S2$ train split.}
    \label{fig_all_ablation}
\end{table}

\begin{itemize}
    \item \textbf{Pose estimation.}
    We investigate the accuracy of our system as a function of the number of matches that are used for the pose estimation (see Table \ref{fig_all_ablation} a). 
    Similar performance is achieved when using $3$, $4$ and $5$ keypoint matches. This is due to a trade-off between using a low number of matches where more images have at least the given amount of correct matches and using a higher number of matches which allows for preciser poses when the minimum number of correct matches is given. As the number of keypoints is increased to 6 or 7 we note a drop in the $\mathrm{AP}^{\mathrm{mesh}}$ score as the large number of images which have less than 6 correct matches (and which therefore will produce inaccurate poses) outweigh the few images for which poses can be computed more accurately.
    \item \textbf{Shape retrieval.}
    We ablate our system in terms of the number of nearest neighbour CAD model renderings that are retrieved (see Table \ref{fig_all_ablation} b). Note that two renderings of the same CAD model, but rendered from different orientations, are considered two separate retrievals for which keypoint matching and pose estimation is performed independently.
    For the S1 split the class-averaged $\mathrm{AP}^{\mathrm{mesh}}$ score for just the top 1 retrieval is 31.1 which is already high. It can be further improved by increasing the number of nearest neighbours that are considered. This effect is even more pronounced on the S2 split where the class-averaged $\mathrm{AP}^{\mathrm{mesh}}$ score increases from 6.9 for top 1 retrieval to 19.5 for top 20 retrieval. The reason for this is that the worse segmentation mask quality on the S2 split compared to the S1 split leads to worse retrieval results (24\% accuracy compared to 65 \% accuracy). At this low retrieval accuracy the benefits of considering additional retrieved shapes is larger compared to the case when the retrieval accuracy is already high.
    \item \textbf{CAD model database.}
    In order to investigate the dependency of the system on the available database size we vary the number of CAD models to which the network has access at test time (see Table  \ref{fig_all_ablation} c). Results on the table, bookcase or wardrobe class demonstrate that when performing shape adaptation even extremely small sets of CAD models can be used to estimate object shapes. Note that the total number of CAD models in the Pix3D \cite{pix3D} dataset is very small. For realistic settings one can easily use hundreds or thousands of CAD models per category therefore allowing for even more precise object shape predictions.
    
    \item \textbf{Pose selection.}
    Finally, we compare the accuracy of the poses when selecting poses based on the minimum distance of reprojected keypoints compared to the approximated silhouette overlap. We obtain an $\mathrm{AP}^{\mathrm{mesh}}$ score of 17.1 (compared to 37.8) on the S1 split and 8.1 (compared to 17.1) on the S2 split. This demonstrates the need for using the estimated silhouette overlap for pose prediction. 
 
\end{itemize}

\section{Estimating the Silhouette Overlap}
\begin{table}[t]
    \centering
    \includegraphics[width=1.0\linewidth]{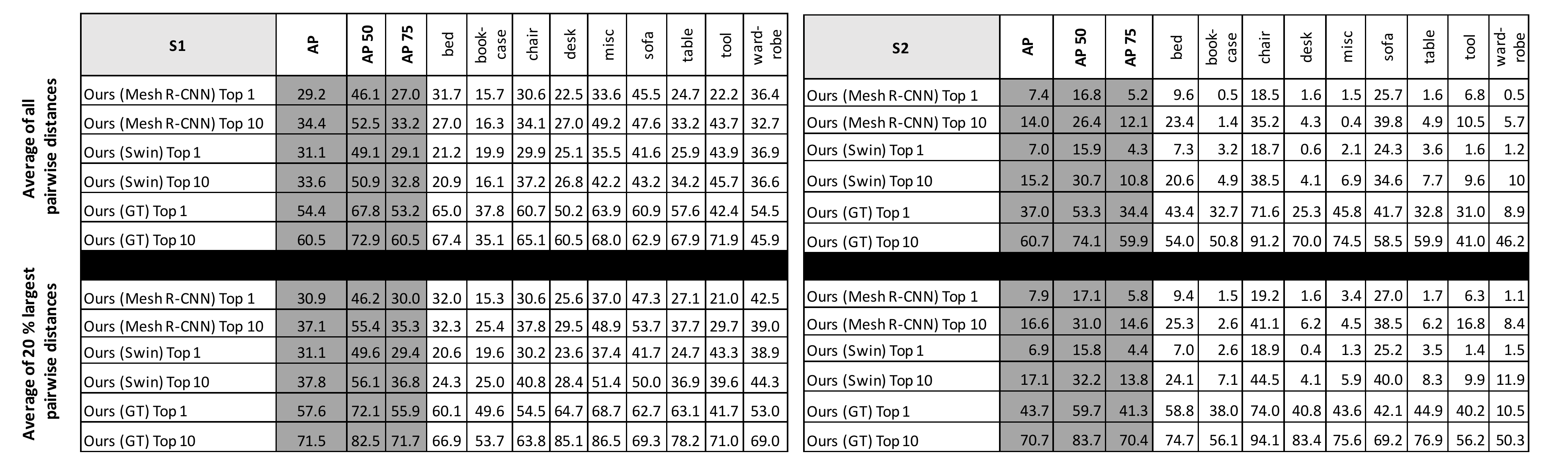}
    \caption{We compare the results we obtain when using different metrics for selecting the final pose. For the results in the top halves of the tables we compute the average over all pairwise distances used to approximate the silhouette overlap of the reprojected CAD model with the predicted segmentation mask. For the results in the lower halves we compute the average over only the 20\% furthest pairwise distances. This is a stronger signal for a good pose as it weighs object outlines (along which pairwise distances are large for bad poses) more compared to area overlaps.}
    \label{appendix_table_comp}
\end{table}

In order to estimate object poses all possible quadruplets of keypoint matches are sampled and their corresponding poses are computed. Choosing the final pose based on the minimum distance of reprojected keypoints often results in sub-optimal poses (see Section \ref{sec_ablations}). Pixel-level inaccuracies of the matches can lead to poses which minimise the reprojection error of the keypoints, but which are very inaccurate. To avoid selecting these poses, poses are chosen based on the estimated silhouette overlap of the reprojected CAD model and the predicted segmentation mask. For this purpose 1000 points are sampled from the retrieved CAD model and reprojected for the current pose estimate. These points are compared to 1000 points sampled inside the predicted segmentation mask by computing the minimum pairwise distance from the reprojected point to points sampled in the segmentation mask and vice versa. Rather than taking the average of all pairwise distances we found it beneficial to only take the average of the 20\% of the largest distances. This poses a stronger signal for matching object outlines and leads on average to more accurate poses (see Table \ref{appendix_table_comp}).

\section{Similarity of Train and Test CAD Models}

We evaluate our proposed model on the Pix3D \cite{pix3D} dataset for which \cite{meshrcnn} introduced two data splits. For the S1 split the 10,069 images are randomly split into 7539 train images and 2530 test images. Under this split all CAD models are seen during training. For the S2 split train and test images are split such that the test images contain CAD models that were not present in the training images. The challenge is therefore to construct a system that given an input image is able to retrieve an unseen CAD model and precisely predict its pose.
\begin{figure*}[t]
    \centering
    \includegraphics[width=1.0\linewidth]{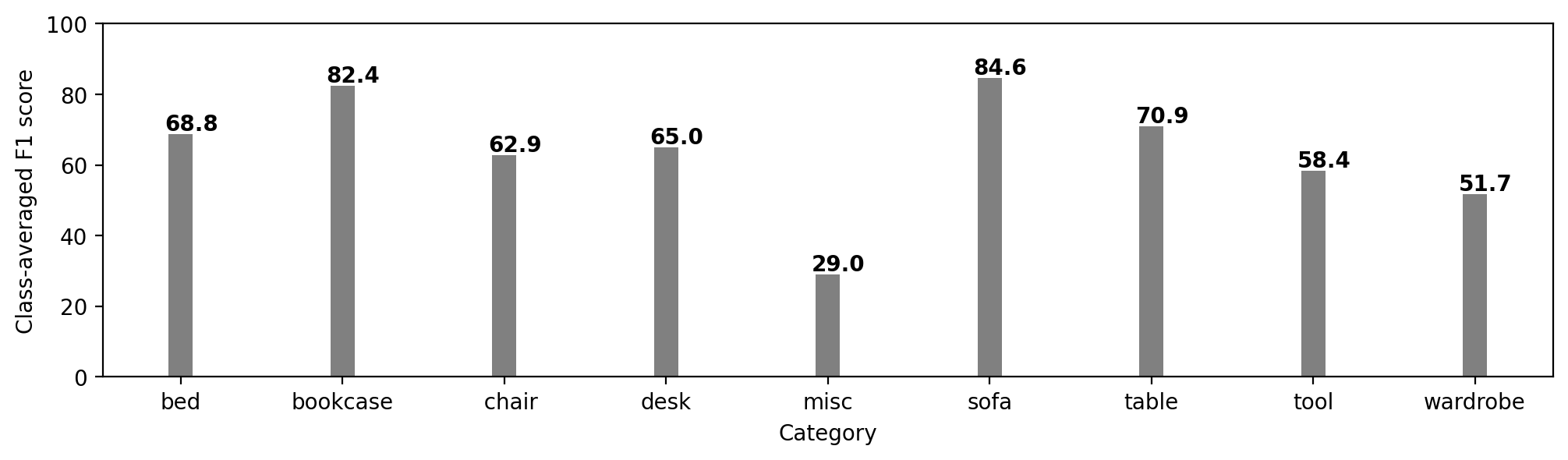}
    \caption{Grey bars provide an indicator for the similarity between CAD models seen during training and unseen CAD models used for testing under the S2 split of Pix3d \cite{pix3D}. Quantitatively grey bars show the class average when the F1 score is computed between every unseen CAD model and its closest matching CAD model (in terms of the F1 score) from the seen ones.}
    \label{S1_s2_supp}
\end{figure*}

\begin{figure*}[t]
    \centering
    \includegraphics[width=1.0\linewidth]{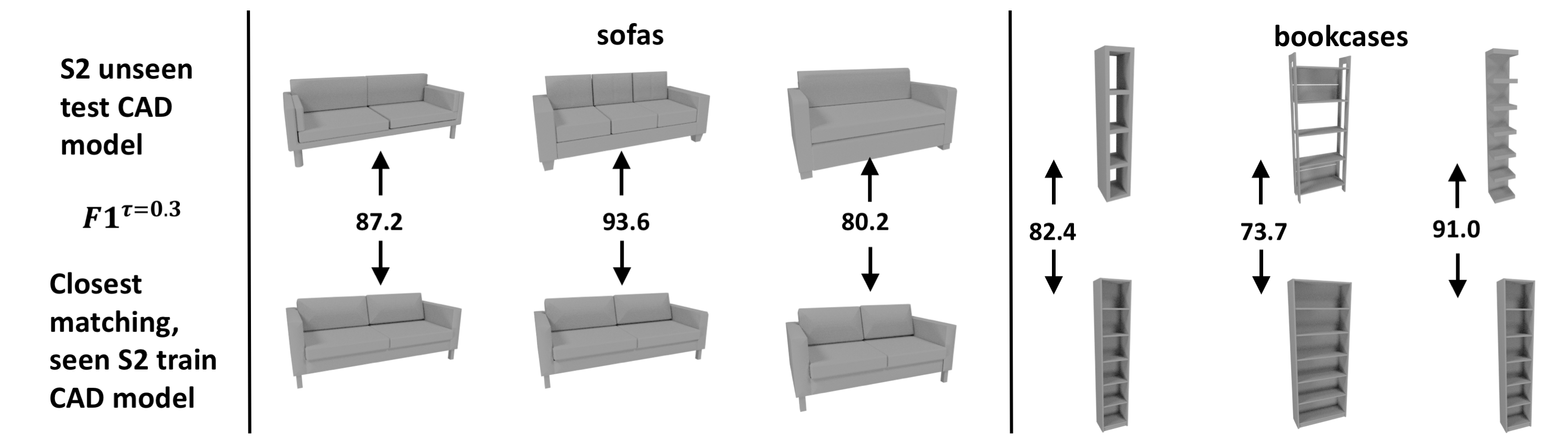}
    \caption{Visualisation of selected test CAD models and their closest matching train CAD models in terms of the F1 score. We note the strong similarity (both visually and in terms of the F1 score) between sofas in the test split and sofas from the train split. While bookcases in the test split also have close matching CAD models in the train split in terms of the F1 score, they differ significantly in their visual appearance. This increases the difficulty for retrieval at test time and explains the poor performance of \cite{mask2cad} on bookcases compared to sofas.}
    \label{seen_unseen_visual}
\end{figure*}

We have demonstrated in our main work that the geometric approach that we follow is more accurate compared to directly predicting object poses \cite{mask2cad} on the S2 split of Pix3D \cite{pix3D}. Further, we will show here that the good performance \cite{mask2cad} achieves on sofas does not require it to retrieve unseen CAD models as for every unseen CAD model in the test images there is a closely fitting CAD model among the seen training CAD models. We quantify this by computing the F1 score at $\tau = 0.3$ between unseen test CAD models and their closest matching CAD models (in terms of F1 score) from the seen ones. We perform this calculation for all unseen CAD models and compute the mean to obtain class averages. These are plotted in gray in Figure \ref{S1_s2_supp}. Note here that the averaged best-possible F1 score for sofas is $84.6$ which is exceptionally high compared to other class averages. This strong similarity (see Figure \ref{seen_unseen_visual} for selected test CAD models and their closest-matching CAD models from the train set) allows \cite{mask2cad} to make accurate shape predictions without retrieving unseen CAD models. We also note that \cite{mask2cad} performs poorly on bookcases despite a high class-averaged F1 score between test CAD models and their closest matching train CAD models. The reason for this is that while good candidate CAD models exist in the seen train CAD models in terms of the F1 score, their different visual appearance (see right side Figure \ref{seen_unseen_visual}) makes them difficult to retrieve at test time.

\section{Qualitative Results in Supplementary Video}
We include further qualitative results of our work in a supplementary video (see \url{https://www.youtube.com/watch?v=n6WD9km8zRQ}). The first sequence of results shows predictions on the Pix3D \cite{pix3D} S1 split, both when using predicted masks as well as ground truth masks. The second sequence contains predictions for the S2 split when access to ground truth CAD models is provided at test time. For both of these cases we perform only pose estimation but no shape adaptation as ground truth CAD models are available. In the third sequence we remove access to ground truth CAD models at test time, only allowing access to S2 train CAD models. We then perform shape adaptation and visualise the shape evolution as a function of iterations of the L-BFGS \cite{lbfgs} minimiser. Note the significantly different shapes to which we can adapt using both predicted masks in the third sequence and ground truth masks in the fourth sequence.

\bibliography{supplementary_material}